\documentclass[sigconf]{acmart}
\AtBeginDocument{%
  \providecommand\BibTeX{{%
    \normalfont B\kern-0.5em{\scshape i\kern-0.25em b}\kern-0.8em\TeX}}}

\copyrightyear{2021} 
\acmYear{2021} 
\setcopyright{acmcopyright}\acmConference[CIKM '21]{Proceedings of the 30th ACM International Conference on Information and Knowledge Management}{November 1--5, 2021}{Virtual Event, QLD, Australia}
\acmBooktitle{Proceedings of the 30th ACM International Conference on Information and Knowledge Management (CIKM '21), November 1--5, 2021, Virtual Event, QLD, Australia}
\acmPrice{15.00}
\acmDOI{10.1145/3459637.3482092}
\acmISBN{978-1-4503-8446-9/21/11}

\usepackage{multicol}
\usepackage{multirow}
\usepackage{graphicx}
\usepackage{adjustbox}
\usepackage{subfig}
\usepackage{xspace}

\newcommand{\modelname}{\textsf{DSKReG}\xspace}

\begin{document}
\fancyhead{}

\title{DSKReG: Differentiable Sampling on Knowledge Graph for Recommendation with Relational GNN}

\author{Yu Wang, Zhiwei Liu, Ziwei~Fan}
\email{{ywang617,zliu213,zfan20}@uic.edu}
\affiliation{
  \institution{University of Illinois at Chicago}
  \country{USA}
}

\author{Lichao Sun}
\email{lis221@lehigh.edu}
\affiliation{
  \institution{Lehigh University}
  \country{USA}
}

\author{Philip S. Yu}
\email{psyu@uic.edu}
\affiliation{
  \institution{University of Illinois at Chicago}
  \country{USA}
}

\begin{abstract}
In the information explosion era, recommender systems (RSs) are widely studied and applied to discover
user-preferred information. A RS performs poorly when suffering from the cold-start issue, which can be alleviated if incorporating Knowledge Graphs~(KGs) as side information.  
However, most existing works neglect the facts that node degrees in KGs are skewed and massive amount of interactions in KGs are recommendation-irrelevant. To address these problems, in this paper, we propose \textbf{D}ifferentiable \textbf{S}ampling on \textbf{K}nowledge Graph for \textbf{Re}commendation with Relational \textbf{G}NN (\modelname) that learns the relevance distribution of connected items from KGs and samples suitable items for recommendation following this distribution. We devise a differentiable sampling strategy, which enables the selection of relevant items to be jointly optimized with the model training procedure. The experimental results demonstrate that our model outperforms state-of-the-art KG-based recommender systems.
The code is available online at \url{https://github.com/YuWang-1024/DSKReG}.
\end{abstract}



\maketitle

\section{Introduction}

Recommender systems have become essential tools for Internet applications to discover potential users' interests \cite{wang2018billion, yang2021consisrec, fan2021continuous, fan2021Modeling, 10.1145/3331184.3331329}.
The crucial part in a recommender system is to characterize collaborative signals from user-item interactions, and recommend similar users with correlated items \cite{he2017neural,wang2019neural,liu2019jscn}.  
However, only leveraging user-item interactions spoils recommendation performance when the data suffers cold-start issues~\cite{wang2019kgat,liu2021augmenting}. 
Therefore, existing works~\cite{wang2019kgat,wang2019KGNN-LS, ai2018cfkg} propose to incorporate knowledge graphs (KGs) as side information~\cite{liu2020kg}, which afford additional semantics among items through intermediate entities, thus alleviating cold-start issues from item perspectives.
\begin{figure*}[t]
    \centering
    \includegraphics[width =6.5in]{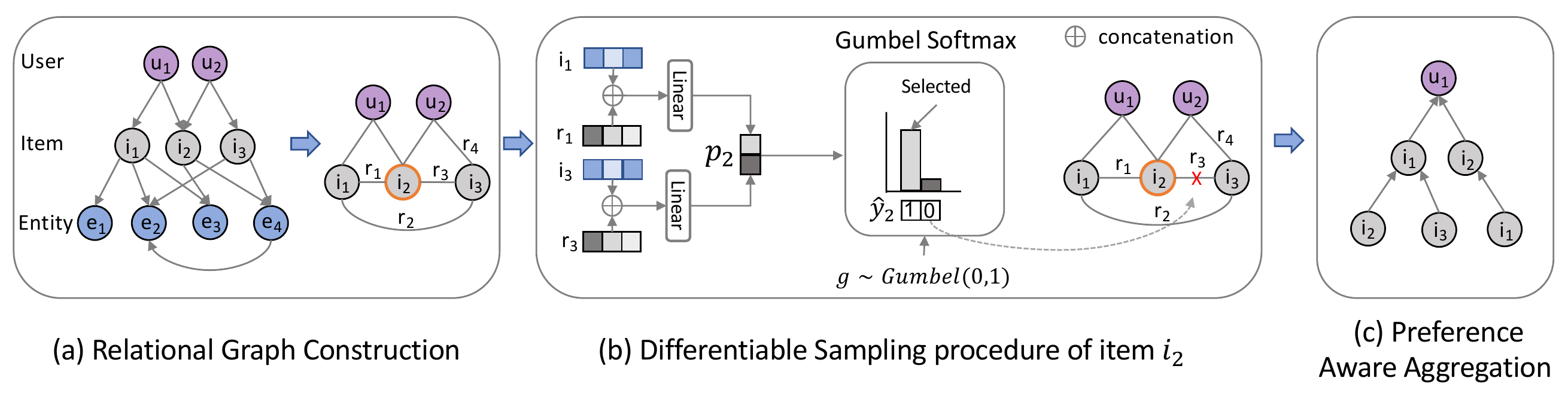}
    \caption{A toy example reflecting the framework of \modelname.    
    a) We construct user-item graph according to users' collaborative interactions, and construct item-item graph by connecting high-order neighbor items. b) For the item $i_2$, we compute the relevance score vector $p_2$ that consists of scores of neighbor items $i_1$ and $i_3$. Afterwards, we apply Gumbel-Softmax over $p_2$ to obtain an approximated one-hot encoder $\hat{y}_2$. The value of $\hat{y}_2$ indicates that neighbor item $i_3$ should be ignored. c) We apply attentive aggregation on relational graph guided by users' preferences. 
    }
    \label{fig:framework}
\end{figure*}

Leveraging the information from KGs requires the model aggregating relevant interactions among items for recommendation~\cite{feng2020atbrg,wang2019kgat}. 
The successes of Graph Neural Networks (GNNs)~\cite{kipf2016gcn,velivckovic2017gat, hamilton2017inductive,liu2020basconv, wang2021explicit} inspire the community designing novel methods for information aggregation.
KGCN~\cite{wang2019KGCN} is one of the pioneering work that adopts GCN~\cite{kipf2016gcn} layers to aggregate entities in KGs to infer item embeddings.
KGNN-LS~\cite{wang2019KGNN-LS} further extends  this idea by assigning user-specific scores on interactions, thus characterizing personalized interests.
KGAT~\cite{wang2019kgat} employs graph attention layers~\cite{velivckovic2017gat} to aggregate both item and user embeddings from KGs. 
Moreover, ATBRG~\cite{feng2020atbrg} constructs sub-graphs from KGs and proposes a relation-aware graph attention layer to adaptively search relevant interactions. 

Despite the effectiveness of existing methods, two limitations are still under-explored: 1) node degree skewness and 2) noisy interactions. Node degree skewness refers that the number of edges for nodes in KGs exhibits a power-law distribution~\cite{clauset2009power}. On the one hand, due to large number of nodes with low degrees and insufficient neighbors, a multi-layer GNN aggregation is required to receive high-order information~\cite{wang2019kgat}. On the other hand, high-order aggregation for nodes with high degrees leads to exponential growth of their receptive field~\cite{xu2018representation}, thus suffering the over-smoothing problem~\cite{li2018deeper,rong2019dropedge}. The second limitation results from massive amount of recommendation-irrelevant interactions in KGs~\cite{feng2020atbrg}. Existing methods infer item embeddings by aggregating all connected entities in KGs. However, directly aggregating those irrelevant entities has no contributions to the representation learning and even increases computational costs, which degrades the performance. 

To address the limitations above, we propose a sampling-based relational GNN to extract recommendation-relevant information from KGs.  
First, we connect items in KGs according to their intermediate entities and create new relations, such as creating a \textit{co-director} relation if two movies are connected by a common director. We illustrate this process in Figure~\ref{fig:framework}(a). This relational graph construction is inspired by works in heterogeneous graph~\cite{wang2019heterogeneous,wei2018unsupervised}. In this way, we could explicitly reveal the item relationships. Second, we adopt sampling-based aggregation of neighbors to avoid the exponential growth of neighbor size, thus alleviating the over-smoothing issue.

However, it is rather challenging to devise a suitable sampling method. Most sampling-based GNNs in KG-based recommendation employ uniform sampling~\cite{wang2019KGNN-LS,wang2019KGCN,wang2018ripplenet} of neighbors, which is unable to distinguish recommendation-relevant relations. 
Moreover, existing sampling strategies~\cite{hamilton2017inductive,zou2019layer,zeng2019graphsaint,wang2020KGPolicy} are independent of the optimization process, which further hinder the end-to-end training manner. RippleNet~\cite{wang2018ripplenet} detachedly samples a fixed-size set of neighbors to infer item embeddings. KGPolicy~\cite{wang2020KGPolicy} employs a disjoint reinforcement learning agents to discover high-quality negative examples in KGs. 
Both methods separate the sampling procedure from the training phase, resulting in a sub-optimal selection of neighbors.

Therefore, we propose a novel model, \textbf{D}ifferentiable \textbf{S}ampling on \textbf{K}nowledge Graph for \textbf{Re}commendation with Relational \textbf{G}NN (\modelname). Given an item, we first compute relevance scores of connected items conditioned on their associated relations and node embeddings. Relevance scores are used to sample top-$K$ relevant neighbor items.
As such, our model can distinguish the recommendation-relevant items among connected neighbors according to relation and item types.
We also adopt Gumbel-Softmax reparameterization trick~\cite{jang2016categorical, xie2019reparameterizable} into the sampling procedure, which approximates the sampling probability from a categorical distribution, thus enabling the sampling procedure to be differentiable. 
Therefore, the sampling component is optimized jointly with the training objective, thus enjoying an end-to-end fashion.

Our contributions are summarized as follows:
1) We compute relevance scores according to relation and item types for sampling, which can navigate model to select recommendation-relevant items. 2) We devise a differentiable sampling strategy to enable the model to refine the sampling procedure jointly with the model optimization.  3) We conduct experiments on three public datasets, and demonstrate the effectiveness of our model. 

\section{Our Approach}

In this section, we first formulate the problem of knowledge-aware recommendation. Then, we propose the \modelname framework, which is shown in Figure~\ref{fig:framework}.

\subsection{Problem Definition}
The objective of knowledge-aware recommendation is to predict whether user $u$ has interest in item $v$ given historical interactions and the KG. Formally, the historical interactions from a set of users $\mathcal{U}$ with the set of items $\mathcal{V}$ are represented as a user-item bipartite graph $\mathcal{G}_{\mathbf{Y}} = \{(u, y_{uv}, v)|u\in \mathcal{U}, v\in \mathcal{V}\}$, where $y_{uv} = 1$ denotes that the user $u$ is interacted with the item $v$ through clicking, purchasing, and \textit{etc}. 
The KG consists of item related properties, such as genres, directors, and casts for movies.
We format the KG as a directed heterogeneous graph $\mathcal{G}_{\mathbf{K}} = \{(h,r,t)|h,t \in \mathcal{E}, r\in \mathcal{R}\}$, such as \textit{(James Cameron, isdirectorof, Titanic)}, where $\mathcal{E}$ and $\mathcal{R}$ denote the set of entities and relations respectively.
Thus, the knowledge-aware recommendation task can be formalized as follows:
\begin{equation}
    \hat{y}_{uv} = \mathcal{F}(u, v|\Theta, \mathcal{G}_{\mathbf{Y}}, \mathcal{G}_{\mathbf{K}}),
\end{equation} where $\hat{y}_{uv}$ is the prediction of user's interest in item $v$, and $\mathcal{F}$ is the learned prediction function with weights $\Theta$.

\subsection{Relational Neighborhood Construction }
The node degree skewness limits the pool of available neighbor items for items with scarce connections in a KG. We propose ``co-interact'' patterns to build up higher order item-item relationships for shortening the path distance between correlated items. Intuitively, those co-interact patterns are important for the recommendation. For example, a user might be interested in books written by the same author.
We extract co-interact patterns from input KG $\mathcal{G}_{\mathbf{K}}$ and construct an item-item co-interact undirected graph $\mathcal{G}_{\mathbf{co}}$ with a new set of co-relations, which is defined as follows:
\begin{equation}
    \mathcal{G}_{\mathbf{co}} = \{(i_1,r',i_2)|\;\text{if}\; (i_1, r, t)\in \mathcal{G}_{\mathbf{K}}\;\text{and}\;(i_2, r, t)\in \mathcal{G}_{\mathbf{K}}\},
\end{equation}
where $r'$ denotes the new ``co-$r$'' relationship.
Following the navigation of these relations, we connect items that have co-interact patterns and construct the item-item graph as shown in Figure~\ref{fig:framework}(a). In this way, we can connect high-order neighbors directly and avoid exponential growth of the receptive field. 
We unify both user-item bipartite graph $\mathcal{G}_{\mathbf{Y}}$ and item-item co-interact graph $\mathcal{G}_{\mathbf{co}}$ into one single graph denoted as relational graph. 
Thus, we can consider all these relations between users and items for subsequent tasks.

\subsection{Differentiable Sampling}
Here, we introduce the proposed differentiable sampling for neighbors selection. We only illustrate it from item's perspective because it is the same process for users. The relevance of co-interact relationships to recommendation varies across users. For example, same genre has more impacts than co-director.
Moreover, co-interact relationships are imbalanced. For example, item-item pairs of co-director are much less than the ones of the same category. This brings up an issue that highly relevant neighbors diminish when the pool of potential neighbors is large. The uniform sampling technique adopted by existing works~\cite{wang2019KGCN, wang2019KGNN-LS} still fails to tackle this issue.
In order to filter out the noise and retain the truly relevant information, we introduce the relation-aware sampling method that assigns weights from relation perspective, as shown in Figure~\ref{fig:framework}(b). The sampling procedure first defines a novel relation-aware relevance score distribution for each item and then samples from it. The relation-aware relevance score distribution of an item $i$ on its co-related neighbors $\mathcal{N}(i)$ is defined as follows:
\begin{equation}
    \label{eq:relevance_dist}
    p(v_{i,j}=1|\mathbf{w},b) = \frac{\exp(\mathbf{w}[\mathbf{r}_{ij}||\mathbf{e}_j]+b)}{\sum_{m \in \mathcal{N}(i)} \exp(\mathbf{w}[\mathbf{r}_{im}||\mathbf{e}_m]+b)},
\end{equation}
where $p(v_{i,j}=1|\mathbf{w},b)$ denotes the plausibility of item $j$ being relevant to the target item $i$;
$\mathbf{w}\in \mathbb{R}^d$ and $b \in \mathbb{R}$ are the learnable weight and bias; $\mathbf{r}_{ij}\in \mathbb{R}^d$ and $\mathbf{e}_j\in \mathbb{R}^d$ are embeddings of relation and neighbor item respectively, and $d$ is the dimension of embeddings. 
The co-relation and neighbor item together determine its neighbor relevance probability, which emphasizes the necessity of relation-awareness in relevance calculation for the sampling.
We apply the same relevance calculation process to users.

Given the calculated relevance distribution, we thus only select top-$K$ most relevant items. Selection procedure of previous works~\cite{feng2020atbrg} is independent of optimization. In other words, the recommendation performance is highly contingent on the result of selection procedure. To make this procedure differentiable and joint with optimization process, we apply the Gumbel-Softmax reparameterization trick. 
Given a Gumbel noise $\mathbf{g}\sim Gumbel(0,1)$, we can draw a soft categorical sample with the following equation:
\begin{equation}
    \mathbf{\hat{y}_i} = \mathrm{Softmax}((\log(\mathbf{p}_i)) + \mathbf{g})/\tau),
\end{equation}
where $\mathbf{p}_i \in \mathbb{R}^d$ consists of relevance score $p(v_{i,j})$ for all the neighbors $j\in\mathcal{N}(i)$ defined in Eq.~(\ref{eq:relevance_dist}), 
and $\tau$ is the annealing temperature. It has been proved~\cite{jang2016categorical, xie2019reparameterizable} that $\mathbf{\hat{y}_i}$ is approximate to a one-hot encoder as $\tau$ goes to $0$. 
We repeat the above procedure for $K$ times and sum the approximated one-hot encoders. 
At each time, the relevance score in $\mathbf{p}_i$ of selected items will be set as $0$.
In this way, we can obtain a $K$-hot vector representing the top-$K$ relevant items selected for subsequent learning procedures.

\subsection{Preference Aware Aggregation}
Besides the factor of relations, we should also consider user preference in the top-$K$ neighbor messages propagation process. As users might have different preferences towards various relations, we take the relations into account in the aggregation.
The aggregation procedure, as shown in Figure~\ref{fig:framework}(c), infers the embedding of item $i$ as follows:
\begin{equation}
\begin{aligned}
     \hat{\mathbf{e}}_i &= \sigma(\mathbf{W}(\mathbf{e}_i + \sum_{j\in \mathcal{N}(i)} \phi_{ij} \mathbf{e}_j)+b),\\
     \phi_{ij} &= \frac{a_{ij} \exp(<\mathbf{e}_{u},\mathbf{r}_{ij}>)}{\sum_{m\in \mathcal{N}(i)}\exp(<\mathbf{e}_{u},\mathbf{r}_{im}>)},
\end{aligned}
\end{equation} 
where $a_{ij}$ is the $j$-{th} position value in the $K$-hot vector of the item $i$ obtained from sampling procedure, 
which indicates whether the item $j$ is selected as a neighbor of item $i$. The $\mathbf{e}_{u}\in\mathbb{R}^d$ is user's embedding. 
For users, we obtain the inferred user embedding $\hat{e}_{u}$ in a similar procedure, but the attentions are calculated using the connected item embeddings.

\subsection{Prediction and Optimization}
We use the dot-product to generate the preference score of user $u$ to item $i$ with the inferred user\slash item embeddings $\hat{\mathbf{e}}_u$ and $\hat{\mathbf{e}}_i$, respectively. The prediction is calculated as follows:
\begin{equation}
    \hat{y}_{ui} = \sigma(\hat{\mathbf{e}}_u^{\intercal}\hat{\mathbf{e}}_i).
\end{equation}
We use the pairwise BPR loss~\cite{rendle2012bpr} to optimize top-$N$ recommendation, which is defined as follows:
\begin{equation}
    \mathcal{L}_{bpr} = \sum_{(u,i,j)\in \mathcal{D}} -\text{log}\sigma\left(\hat{y}(u,i) - \hat{y}(u,j)\right) + \lambda||\Theta||_2^2,
\end{equation}
where $\mathcal{D}$ is a set of triplets, each of them is composed of user $u$, an interacted item $i$ and one sampled negative item from items that user $u$ never interacts with.

\begin{table*}[]
    \caption{Overall Comparison}
    \label{tab:overall}
    \small
    \centering
    \begin{tabular}{l|l|c|c|c|c|c|c|c|c|c}
         \hline
         &&\multicolumn{3}{c|}{Recall}   &\multicolumn{3}{|c|}{Precision} &\multicolumn{3}{|c}{NDCG}\\
         \hline
         Dataset&Model&R@5&R@10&R@20&P@5&P@10&P@20&N@5&N@10&N@20\\
         \hline
         \multirow{6}{*}{Last.FM}&CFKG&	0.0028&	0.0069	&0.0069	&  0.0011&	0.0011&	0.0006	&  0.0024&	0.0042&	0.0042	 \\
         &KGAT& 0.0661& 0.0996& 0.1483& 0.0209& 0.0174& 0.0131&\textbf{ 0.0627}& \underline{0.0781} &\underline{0.0951}\\
         &KGNN-LS& \underline{0.0801} &\underline{0.1297}& \underline{0.2073}&\underline{0.0275} &\underline{0.0229}& \underline{0.0178}&0.0518 &0.0705 &0.0929\\
         &RippleNet&0.0762 &0.0962 &0.1313&0.0253 &0.0163 &0.0114 &0.0595& 0.0672 &0.0776\\
         &\modelname&\textbf{0.0870} &\textbf{0.1481} &\textbf{0.2084}&\textbf{0.0298}& \textbf{0.0253} &\textbf{0.0178}&\underline{0.0621}& \textbf{0.0889}& \textbf{0.1086} \\
         \hline
         & Improvement&8.6\%&14.1\%& 0.5\%&8.4\%&10.5\%&0.0\%&-0.9\% &13.8\%&14.2\%\\
         \hline
                  \multirow{6}{*}{BookCrossing}&CFKG&	0.0403&	0.0403	&0.0406&  0.0156 &   0.0078  &  0.0041 & \underline{0.0661}   & \underline{0.0661}  &  \underline{0.0668} \\
         &KGAT& 0.0059& 0.0141&0.0321& 0.0044& 0.0046& 0.0041& 0.0117& 0.0170& 0.0242\\
         &KGNN-LS&\underline{0.0437} &\underline{0.0528} &\underline{0.0761}& \underline{0.0181}& \underline{0.0119}& \textbf{0.0089}& 0.0425 &0.0455& 0.0519\\
         &RippleNet&0.0387 &0.0457 &0.0596 &0.0157& 0.0097& 0.0070&0.0424 &0.0442 &0.0485 \\
         &\modelname&\textbf{0.0506} &\textbf{0.0593} &\textbf{0.0763}& \textbf{0.0201}& \textbf{0.0125}& \underline{0.0088}&\textbf{0.0722} &\textbf{0.0782}& \textbf{0.0882}\\
         \hline
         &Improvement&15.8\%&12.3\%&0.2\%&11.0\%&5.0\%&-1.1\%&9.2\%&18.3\%&32.0\%\\
         \hline
                  \multirow{6}{*}{MoiveLens-Sub}&CFKG&   0.0013&	0.0057&	0.0086&  0.0017	&0.0014	&0.0010	 & 0.0030&	0.0059&	0.0076	\\
         &KGAT&   0.0205  &  0.0361 &   0.0699&  0.0071  &  0.0060 &   0.0059  & \underline{0.0184} &   0.0254  &  0.0386 \\
         &KGNN-LS &0.0201 &0.0325& 0.0678&  0.0067 &0.0055& 0.0058&0.0154& 0.0205& 0.0313\\
         &RippleNet& \underline{0.0246} &\underline{0.0535}&\textbf{ 0.1097}& \underline{0.0089} &\underline{0.0091}& \underline{0.0087}& 0.0175& \underline{0.0289}& \underline{0.0450}\\
         &\modelname& \textbf{0.0295 }&\textbf{0.0618 }&\underline{0.1086} & \textbf{0.0107} &\textbf{0.0107 }&\textbf{0.0095}&\textbf{0.0275} &\textbf{0.0432 }&\textbf{0.0619}\\
         \hline
         &Improvement& 19.9\%&15.5\%&-1.0\%&20.2\%&17.6\%&9.2\%&49.4\%&49.5\%&37.5\%\\
         \hline
    \end{tabular}
\end{table*}
\section{Experiments}
In this section, we introduce the experimental settings and compare our model with state-of-the-art methods on three common benchmark recommendation datasets. Then, we perform the ablation study and discuss effects of the model's components. 
\subsection{Experimental Settings}
\begin{table*}[t]
    \caption{Dataset Statistics}
    \centering
    \begin{tabular}{lrrrcrcrcr}
    \hline
        Dataset &  \# user &\# item &\# train user& \# test user &\# rating &density&\#entity& \#relation& \# triples\\
         \hline
        Last.FM & 1,872 & 3,846 & 1,872 & 363 & 21,173 &11.3& 9,366 & 60 & 15,518\\
        BookCrossing&17,860&14,967&17,860&497&69,873&3.9 &77,903 &25&151,500\\
        MovieLens-Sub &5,423 & 2,445 & 6,036&563 &37,858&7.0&125,061&12&539,350\\
    \hline
    \end{tabular}

    \label{tab:datasets}
\end{table*}
\textbf{Datasets.} To evaluate the effectiveness of our model, we perform experiments on three benchmark datasets: Last.FM, BookCrossing and MovieLens-Sub. Last.FM is a set of online listening information from Last.fm website. BookCrossing contains users' ratings of books. MovieLens-Sub comes from a widely used benchmark dataset: MovieLens-1M, which contains users' ratings of movies from MovieLens website. 
Table~\ref{tab:datasets} provides detail empirical statistics of these datasets. We define the density of a dataset as the division of the number of ratings by the number of users, which indicates the average number of ratings per user. During the empirical study of these datasets, we found the density of original MovieLens-1M is 62.4, which is extremely large compared to other datasets. To make the dataset fit the cold-start scenario, we only randomly chose $10\%$ of all ratings to construct MovieLens-Sub.\\
\textbf{Baselines.} We compare our model with state-of-the-art methods: KGAT \cite{wang2019kgat}, KGNN-LS \cite{wang2019KGNN-LS}, RippleNet \cite{wang2018ripplenet} and knowledge embedding based method CFKG \cite{ai2018cfkg}.

\subsection{Results}
To evaluate top-$N$ recommendation and preference ranking performance, we use three standard metrics: Recall, Precision, and NDCG. For each dataset, we randomly sample a subset of users for evaluation. Then, we rank the users' preference scores over all items except training items. Finally, we compute the Recall, Precision, and NDCG on top 5, top 10, and top 20 items, respectively.

As Table~\ref{tab:overall} shows, our model outperforms state-of-the-art methods significantly in most cases. Compared to the strongest baseline model, we manage to improve the performance by 7.73\%, 6.2\%, and 9.03\% on Last.FM on average for Recall, Precision, and NDCG respectively. 
Similarly, we outperform the best baseline model by 9.43\%, 4.97\%, and 19.83\% on BookCrossing. On the MovieLens-Sub dataset, we improve the performance by 11.47\%, 15.60\%, and 45.47\% respectively. 
These results indicate the effectiveness of our model. Surprisingly, our model improve the NDCG by a significant margin. Specifically, we improve the NDCG@20 by 14.2\%, 32.0\%, and 37.5\% on three datasets respectively. Since NDCG measures the recommendation quality taking position significance and the number of items into account, these results demonstrate the superiority of our model in recommendation.

\begin{figure}[tb]
\centering
\subfloat[Last.FM]{\includegraphics[width=1.1in]{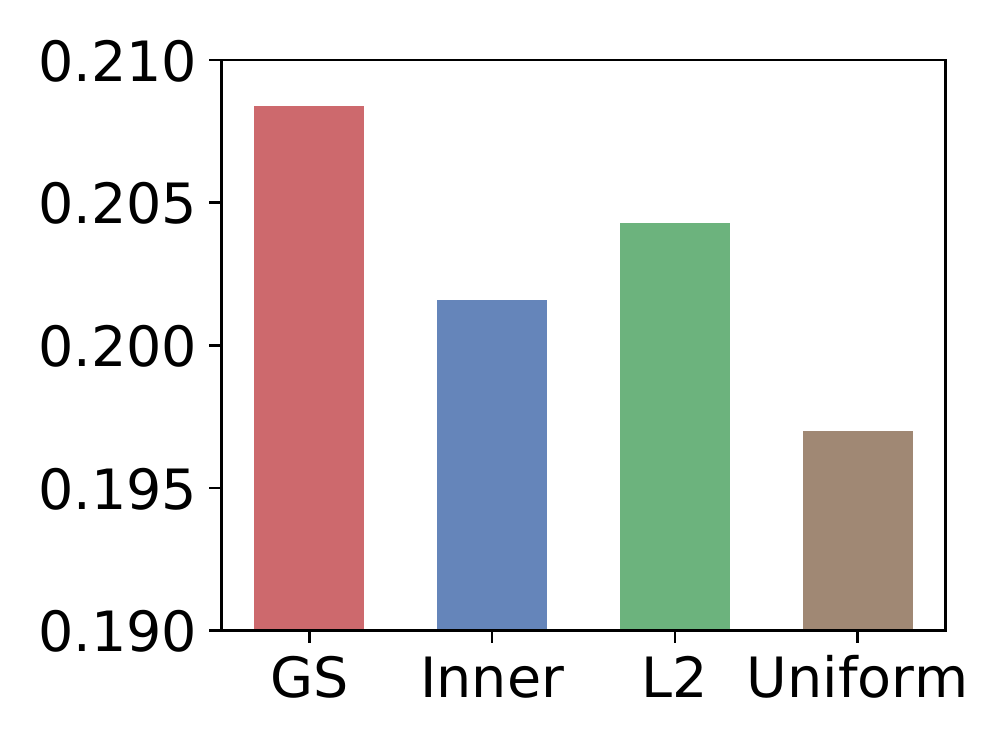}} \subfloat[BookCrossing] {\includegraphics[width=1.1in]{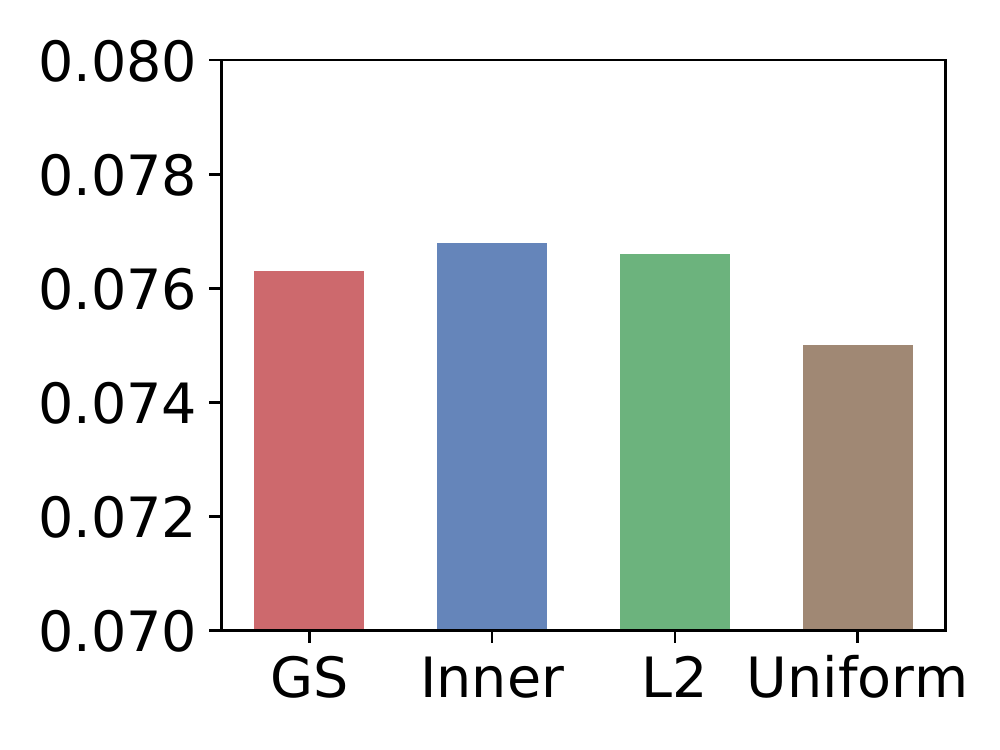}}
\subfloat[MovieLens-Sub]{\includegraphics[width=1.1in]{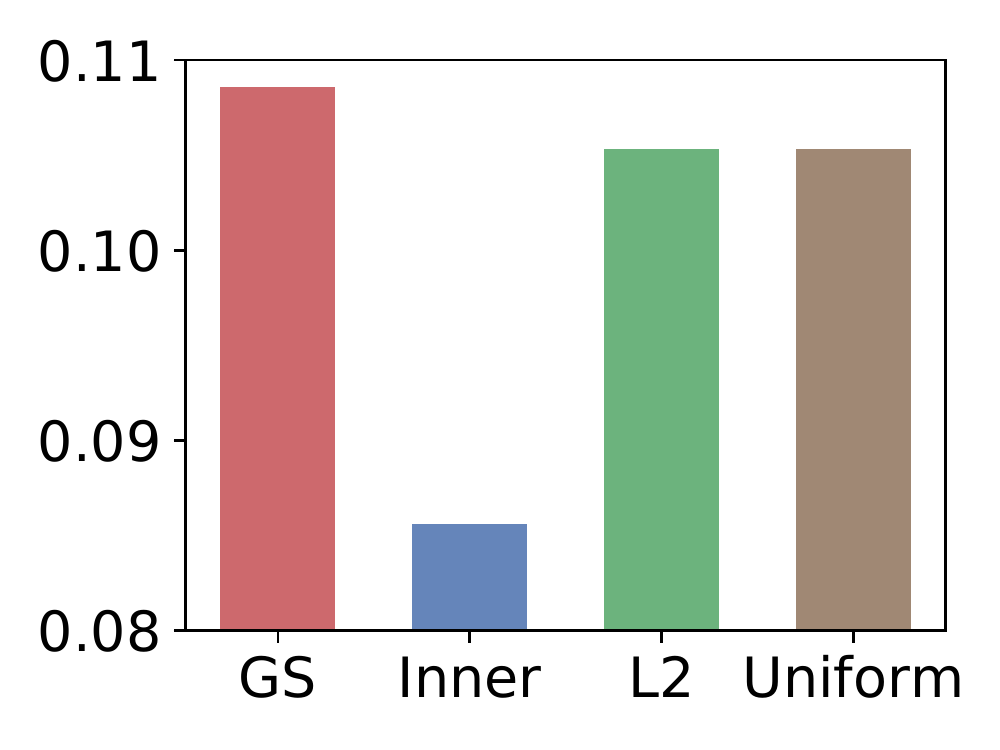}}
\caption{Recall@20 using different sampling strategy} \label{fig:sample}
\end{figure}

\begin{figure}[tb]
\centering
\subfloat[Last.FM]{\includegraphics[width=1.1in
]{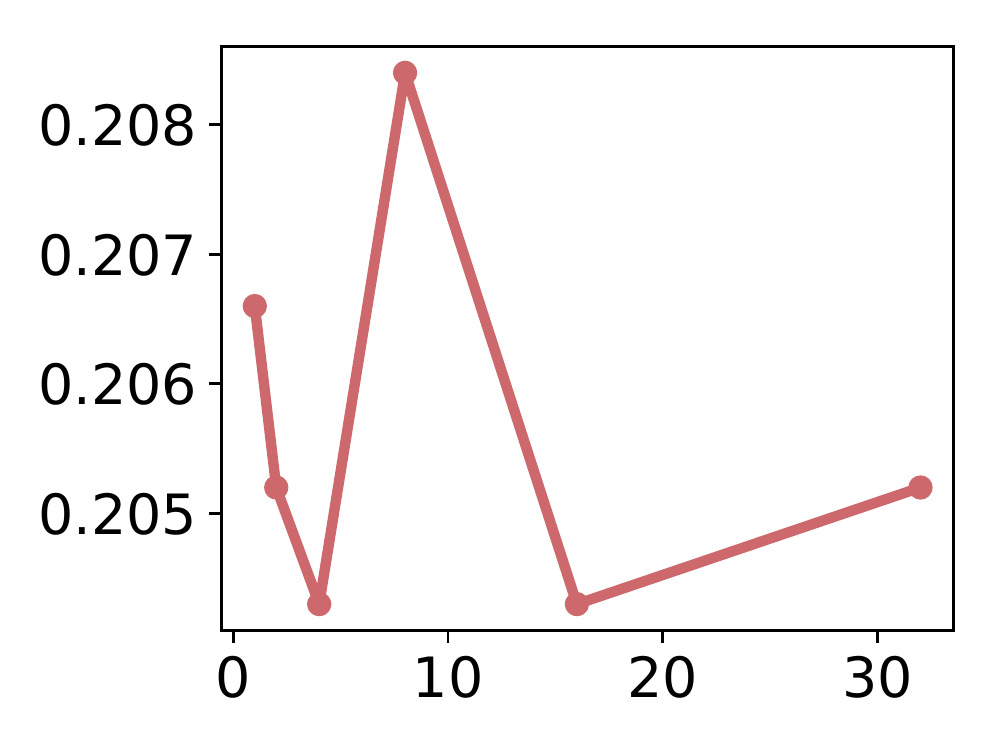}}\hspace{-1mm}
 \subfloat[BookCrossing]{\includegraphics[width=1.1in
 ]{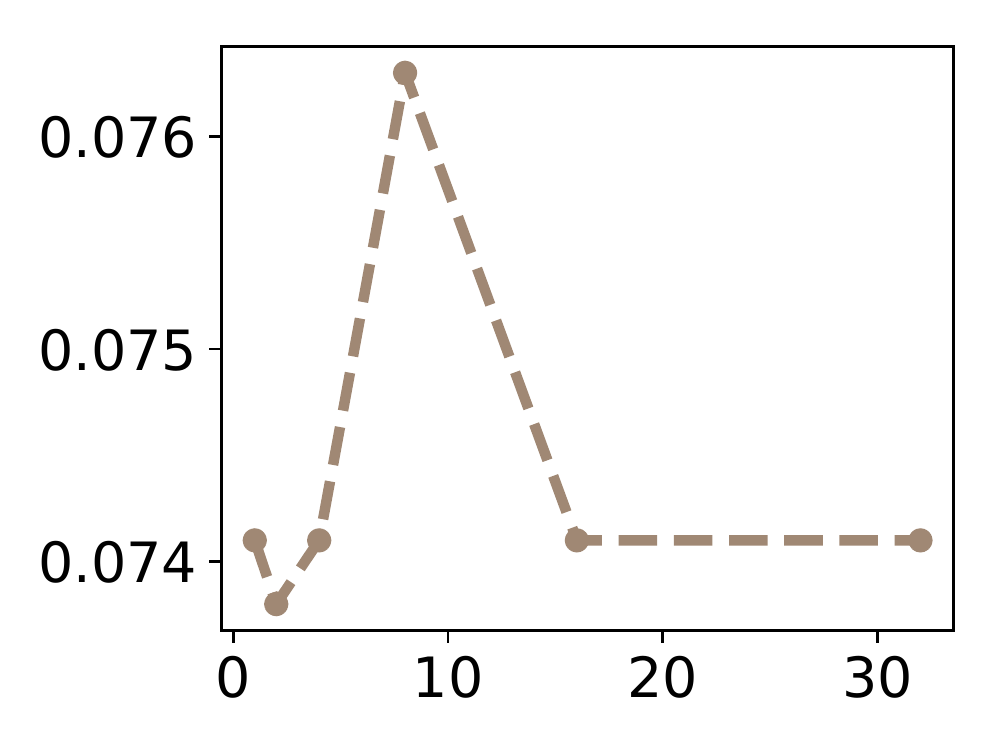}}\hspace{-1mm}
\subfloat[MovieLens-Sub]{\includegraphics[width=1.1in
]{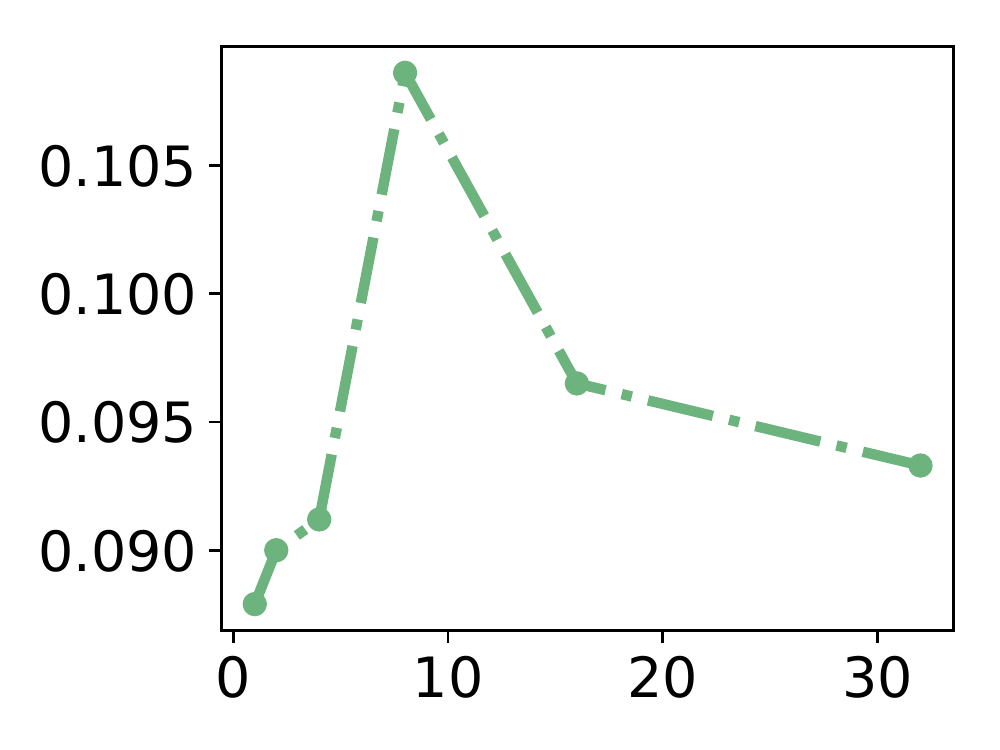}} 
\caption{Recall@20 with different neighbor size} \label{fig:K}
\end{figure}

\subsection{Ablation Study}
In this section, we perform the ablation study to better understand effects of different components of our model.
\paragraph{The Effect of Relation-aware Sampling} To examine the effect of relation-aware sampling, we compare our model with different sampling strategies. As shown in Figure~\ref{fig:sample}, uniform indicates we randomly select $K$ neighbors for each item; L2 means we use the $L2$-norm of difference between relation and neighbor item embeddings as the categorical sampling distribution; Inner represents that we use the inner product between relation and item embeddings as sampling probability; We denote the differentiable sampling method using Gumbel-Softmax as GS. The experimental results indicate that the GS outperforms the others on Last.FM and MovieLens-Sub. On BookCrossing, models using L2 distance and Inner product metrics can achieve comparable results with GS. The possible reason is that relations among items in this dataset are relatively simple. 
As shown in Table~\ref{tab:datasets}, BookCrossing has smaller number of relations in the original KG dataset than that in Last.FM, and smaller number of triples than that in MovieLens-Sub.
The L2 distance and Inner product metric are sufficient to model the item relations. However, in dealing with complex item relations, GS significantly outperforms the other metrics.

\paragraph{The Effect of Sampling Size.} To examine the effectiveness of neighbor size, we perform experiments with different $K$, which is the size of the neighborhood after sampling. As shown in Figure~\ref{fig:K}, the best neighbor size is $8$ for Last.FM, BookCrossing, and MovieLens-Sub. This indicates that only a small portion of items are relevant.

Our model can correctly select this valuable information for aggregation, which enables our model to achieve the best performance with only eight neighbors.

\section{Conclusion}
In this paper, we proposed a novel framework \modelname to alleviate the node degree skewness and noisy interactions limitations when tackling KG-based recommendation. \modelname is a sampling-based relational GNN, which extracts recommendation-relevant information from KGs.
We devised a differentiable sampling strategy for \modelname, which is jointly optimized with the model to learn how to select top-$K$ relevant items for aggregation. 
We conducted experiments on three public dataset to demonstrate the effectiveness of \modelname in improving the recommendation performance. 

\section{ACKNOWLEDGEMENTS}
This work is supported in part by NSF under grants III-1763325, III-1909323,  III-2106758, and SaTC-1930941. 
\bibliographystyle{ACM-Reference-Format}
\balance
\bibliography{reference}
\end{document}